\let\origvec\vec 
\documentclass[a4paper]{llncs}
\usepackage[utf8]{inputenc}
\usepackage[T1]{fontenc}
\usepackage{tabularx}
\usepackage{multirow}
\usepackage{multicol}
\usepackage{longtable}
\usepackage{graphicx}
\usepackage{subfigure}
\usepackage{url}
\usepackage[colorlinks=true,
            linkcolor=black,
            anchorcolor=black,
            citecolor=black,
            filecolor=black,
            menucolor=black,
            runcolor=black,
            urlcolor=black]{hyperref}

\let\vec\origvec 
\usepackage{amsmath}
\usepackage{amssymb}
\usepackage{mathtools}
\usepackage{complexity}

\usepackage{xspace}

\usepackage{paralist} 

\usepackage{color}
\definecolor{darkblue}{rgb}{0,0,.6}
\definecolor{darkred}{rgb}{.6,0,0}
\definecolor{darkgreen}{rgb}{0,.6,0}
\definecolor{red}{rgb}{.98,0,0}

\usepackage{listings}

\usepackage{tikz}
\usetikzlibrary{arrows}
\usetikzlibrary{positioning}

\title{An Evolutionary Algorithm to Learn SPARQL Queries for Source-Target-Pairs}
\subtitle{Finding Patterns for Human Associations in DBpedia}

\author{Jörn Hees \and Rouven Bauer \and Joachim Folz \and Damian Borth \and Andreas Dengel}

\institute{Computer Science Department, University of Kaiserslautern, Germany
\and Knowledge Management Department, DFKI GmbH, Kaiserslautern, Germany
\email{\{firstname.lastname\}\,@dfki.de}
}

\begin{document}
\newcommand{\redline}[1]{\ignorespaces\begin{scriptsize}\begin{itemize}\setlength{\parskip}{0pt}%
	\setlength{\parsep}{0pt}\color{darkred}#1\end{itemize}\end{scriptsize}\ignorespaces}
\newcommand{\redlineDone}[1]{}
\newcommand{\FIXME}[1]{\marginpar{\color{darkred}FIXME: #1}}
\newcommand{\TODO}[1]{\marginpar{\color{darkred}TODO: #1}}

\newcommand{\introduce}[1]{\emph{#1}}
\newcommand{\reuse}[1]{\emph{#1}}

\newcommand{\lO}{\mathcal{O}}

\newcommand{\curiesize}{\scriptsize}
\newcommand{\curie}[2]{\text{\curiesize \href{#1}{\nolinkurl{#2}}}}
\newcommand{\dbr}[1]{\curie{http://dbpedia.org/resource/#1}{dbr:#1}}
\newcommand{\dbprop}[1]{{\curiesize \href{http://dbpedia.org/property/#1}{\nolinkurl{dbprop:#1}}}}
\newcommand{\dbo}[1]{{\curiesize \href{http://dbpedia.org/ontology/#1}{\nolinkurl{dbo:#1}}}}
\newcommand{\eat}[1]{{\curiesize \href{http://www.eat.rl.ac.uk/\##1}{\nolinkurl{eat:#1}}}}
\newcommand{\assoc}[1]{{\curiesize \href{https://w3id.org/associations/vocab\##1}{\nolinkurl{a:#1}}}}
\newcommand{\dbpam}[1]{{\curiesize \href{https://w3id.org/associations/mapping_eat_dbpedia\##1}{\nolinkurl{dbpam:#1}}}}
\newcommand{\gold}[1]{{\curiesize \href{http://purl.org/linguistics/gold/#1}{\nolinkurl{gold:#1}}}}
\newcommand{\rdf}[1]{{\curiesize \href{http://www.w3.org/1999/02/22-rdf-syntax-ns\##1}{\nolinkurl{rdf:#1}}}}
\newcommand{\rdfs}[1]{{\curiesize \href{http://www.w3.org/2000/01/rdf-schema\##1}{\nolinkurl{rdfs:#1}}}}
\newcommand{\owl}[1]{{\curiesize \href{http://www.w3.org/2002/07/owl\##1}{\nolinkurl{owl:#1}}}}
\newcommand{\dcterms}[1]{{\curiesize \href{http://purl.org/dc/terms/#1}{\nolinkurl{dcterms:#1}}}}
\newcommand{\umbel}[1]{{\curiesize \href{http://umbel.org/umbel/rc/#1}{\nolinkurl{umbel:#1}}}}
\newcommand{\skos}[1]{{\curiesize \href{http://www.w3.org/2004/02/skos/core\##1}{\nolinkurl{skos:#1}}}}

\newcommand{\iri}[1]{<\url{#1}>}
\newcommand{\prefix}[2]{\url{#1}: \iri{#2}}
\newcommand{\iripart}[1]{\url{#1}}
\newcommand{\sparql}[1]{\text{\curiesize \texttt{#1}}}
\newcommand{\patterni}[1]{\{\sparql{#1}\}}
\newcommand{\pattern}[1]{\vspace{-1.5ex}\begin{center}\patterni{#1}\end{center}\vspace{-1ex}}
\newcommand{\patternlabel}[2]{\vspace{-1.5ex}\begin{center}#1: \patterni{#2}\end{center}\vspace{-1ex}}
\newcommand{\association}[2]{``#1 - #2''}
\newcommand{\semassociation}[2]{(#1, #2)}
\newcommand{\stimulus}[1]{``#1''}
\newcommand{\response}[1]{``#1''}
\newcommand{\source}{\sparql{?source}\xspace}
\newcommand{\target}{\sparql{?target}\xspace}

\newcommand{\sym}[1]{``#1''}

\renewcommand{\subsubsection}[1]{\vspace{1ex}\noindent\textbf{#1}}

\maketitle

\begin{abstract}
Efficient usage of the knowledge provided by the Linked Data community is often hindered by the need for domain experts to formulate the right SPARQL queries to answer questions.
For new questions they have to decide which datasets are suitable and in which terminology and modelling style to phrase the SPARQL query.

In this work we present an evolutionary algorithm to help with this challenging task.
Given a training list of source-target node-pair examples our algorithm can learn patterns (SPARQL queries) from a SPARQL endpoint.
The learned patterns can be visualised to form the basis for further investigation, or they can be used to predict target nodes for new source nodes.

Amongst others, we apply our algorithm to a dataset of several hundred human associations (such as \association{circle}{square}) to find patterns for them in DBpedia.
We show the scalability of the algorithm by running it against a SPARQL endpoint loaded with $> 7.9$ billion triples.
Further, we use the resulting SPARQL queries to mimic human associations with a Mean Average Precision (MAP) of $39.9 \%$ and a Recall$@10$ of $63.9 \%$.
\end{abstract}

\section{Introduction}\label{sec:intro}
	The Semantic Web~\cite{BernersLeeHL2001SemanticWeb} and its Linked Data~\cite{Bizer2009LinkedDataTheStorySoFar} movement have brought us many great, interlinked and freely available machine readable RDF~\cite{RDFConcepts2004} datasets, often summarized in the Linking Open Data Cloud\footnote{\url{http://lod-cloud.net/}}.
	Being extracted from Wikipedia and spanning many different domains, DBpedia~\cite{Bizer2009DBpedia} forms one of the most central and best interlinked of these datasets.

	Nevertheless, even with all this easily available data, using it is still very challenging:
	For a new question, one needs to know about the available datasets, which ones are best suited to answer the question, know about the way knowledge is modelled inside them and which vocabularies are used, before even attempting to formulate a suitable SPARQL\footnote{\url{https://www.w3.org/TR/rdf-sparql-query/}} query to return the desired information.
	The noise of real world datasets adds even more complexity to this.


	In this paper we present a graph pattern learning algorithm that can help to identify SPARQL queries for a relation $\mathcal{R}$ between node pairs $(s,t) \in \mathcal{R}$ in a given knowledge graph $G$\footnote{For our purpose $G$ is a set of RDF triples, typically accessible via a given SPARQL endpoint.}, where $s$ is a source node and $t$ a target node.
	$\mathcal{R}$ can for example be a simple relation such as ``given a capital $s$ return its country $t$'' $\mathcal{R}_{cc}$ or a complex one such as ``given a stimulus $s$ return a response $t$ that a human would associate'' $\mathcal{R}_{ha}$.

	To learn queries for $\mathcal{R}$ from $G$, without any prior knowledge about the modelling of $\mathcal{R}$ in $G$, we allow users to compile a ground truth set of example source-target-pairs $\mathcal{GT} \subseteq \mathcal{R}$ as input for our algorithm.
	For example, for relation $\mathcal{R}_{cc}$ between capital cities and their countries, the user could generate a ground truth list $\mathcal{GT} = \{$(\dbr{Berlin}, \dbr{Germany}), (\dbr{Paris}, \dbr{France}), (\dbr{Oslo}, \dbr{Norway})$\}$.
	Given $\mathcal{GT}$ and the DBpedia SPARQL endpoint\footnote{\url{http://dbpedia.org/sparql}}, our graph pattern learner then learns a set of graph patterns $gpl(\mathcal{GT}, G)$ such as:
	\patternlabel{$gp_1$}{?source \dbo{country} ?target}
	\patternlabel{$gp_2$}{?target \dbo{capital} ?source. ?target a \dbo{Country}}

	In this paper, a graph pattern $gp \in gpl(\mathcal{GT}, G) \subset GP$ is an instance of the infinite set of SPARQL basic graph patterns\footnote{\url{https://www.w3.org/TR/sparql11-query/\#BasicGraphPatterns}} $GP$.
	Each $gp$ has a corresponding SPARQL ASK and SELECT query.
	We denote their execution against $G$ as $\mathrm{ASK}(gp)$ and $\mathrm{SELECT}(gp)$.
	The graph patterns can contain SPARQL variables, out of which we reserve \source and \target as special ones.
	A mapping $\Phi$ can be used to bind variables in $gp$ before execution.

	The resulting learned patterns can either be inspected or be used to predict targets by selecting all bindings for \target given a source node $s_i$:
	$$\text{prediction}_{gp}(s_i) = \mathop{\mathrm{SELECT}}_{?target}(\phi_{?source \coloneqq s_i}(gp))$$
	For example, given the source node \dbr{London} the pattern $gp_1$ can be used to predict $\dbr{United_Kingdom} \in \text{prediction}_{gp_1}(\dbr{London})$.

	The remainder of this paper is structured as follows:
	We present related work in Section~\ref{sec:relwork}, before describing our graph pattern learner in detail in Section~\ref{sec:gp_learner}.
	In Sections~\ref{sec:visualisation} and \ref{sec:prediction} we will then briefly describe visualisation and prediction techniques before evaluating our approach in Section~\ref{sec:eval}.

\section{Related Work}\label{sec:relwork}
	To the best of our knowledge, our algorithm is the first of its kind.
	It is unique in that it can learn a set of SPARQL graph patterns for a given input list of source-target-pairs directly from a given SPARQL endpoint.
	Additionally, it can cope with scenarios in which there is not a single pattern that covers all source-target-pairs.

	Many other algorithms exist, which learn vector space representations from knowledge graphs.
	An excellent overview of such algorithms can be found in \cite{Nickel2015ReviewRelationalMachineLearningForKnowledgeGraphs}.
	We are however not aware that any of these algorithms have the ability of returning a list of SPARQL graph patterns that cover an input list of source-target-pairs.

	There are other approaches that help formulating SPARQL queries, mostly in an interactive fashion such as RelFinder \cite{Heim2009RelFinder,Heim2010RelFinder} or AutoSPARQL~\cite{Lehmann2011AutoSPARQL}.
	Their focus however lies on finding relationships between a short list of entities (not source-target-pairs) or interactively formulating SPARQL queries for a list of entities of a single kind.
	They cannot deal with entities of different kinds.

	Wrt.\ SPARQL pattern learning, there is an approach for pattern based feature construction~\cite{LawrynowiczPotoniec2014PatternBasedFeatureConstructionSPARQL} that focuses on learning SPARQL patterns to use them as features for binary classification of entities.
	It can answer questions such as: does an entity belong to a predefined class?
	In contrast to that, our approach focuses on learning patterns between a list of source-target-pairs for entity prediction: given a source entity predict target entities.
	To simulate target entity prediction for a single given source with binary classification, one would need to train $n$ classifiers, one for $n$ potential target entities.

	In the context of mining patterns for human associations and Linked Data, we previously focused on collecting datasets of semantic associations directly from humans \cite{Hees2010LDGames,HeesKhamis2013KnowledgeTestGame}, ranking existing facts according to association strengths \cite{Hees2011BetterRelationsKI,Hees2012SeCoBetterRelations} and mapping the Edinburgh Associative Thesaurus \cite{Kiss1973EAT} to DBpedia \cite{Hees2016EAT}.
	None of these previous works directly focused on identifying existing patterns for human associations in existing datasets.

\section{Evolutionary Graph Pattern Learner}\label{sec:gp_learner}
	The outline of our graph pattern learner is similar to the generic outline of evolutionary algorithms:
	It consists of individuals (in our case SPARQL graph patterns $gp_i \in GP$), which are evaluated to calculate their fitness.
	The fitter an individual is, the higher its chance to survive and reach the next generation.
	The individuals of a generation are also referred to as population.
	In each generation there is a chance to mate and mutate for each of the individuals.
	A population can contain the same individual (graph pattern) several times, causing fitter individuals to have a higher chance to mate and mutate over several generations.

	As mentioned in the introduction, the training input of our algorithm is a list of ground truth source-target-pairs $gtp_i = (s_i, t_i) \in \mathcal{GT}$.

	Due to size limitations, we will focus on the most important aspects of our algorithm in the following.
	For further detail please see our website\footnote{\url{https://w3id.org/associations}} where you can find the source-code, visualisation and other complementary material.

\subsection{Coverage}\label{sec:coverage}
	Before describing the realisation of the components of our evolutionary learner, we want to introduce our concept of coverage.

	We say that a graph pattern $gp_i$ covers, models or fulfils a source-target-pair $(s_j, t_j)$ if the evaluation of its SPARQL ASK query returns true:
	$$\mathrm{ASK}(\phi_{\source \coloneqq s_i, \target \coloneqq t_i}(gp))$$

	Our algorithm is not limited to learning a single best pattern for a list of ground truth pairs, but it can learn multiple patterns which together cover the list.

	We realise this by invoking our evolutionary algorithm in several \introduce{runs}.
	In each run a full evolutionary algorithm is executed (with all its generations).
	After each run the resulting patterns are added to a global list of results.
	In the following runs, all ground truth pairs which are already covered by the patterns from previous runs become less rewarding for a newly learnt pattern to cover.
	Over its runs our algorithm will thereby re-focus on the left-overs, which allows us to maximise the coverage of all ground truth pairs with good graph patterns.

\subsection{Fitness}\label{sec:fitness}
	In order to evaluate the fitness of a pattern, we define the following dimensions to capture what makes a pattern ``good''.


	\begin{itemize}
	\item High \introduce{recall}:\\
		A good pattern fulfils as many of the given ground truth pairs $\mathcal{GT}$ as possible:
		\begin{eqnarray*}
			\text{gt\_matches}_{gp} & = & |\{(s_i, t_i) \in \mathcal{GT}|\mathrm{ASK}(\phi_{?source \coloneqq s_i, ?target \coloneqq t_i}(gp))\}| \\
			\text{recall}_{gp} & = & \frac{\text{gt\_matches}_{gp}}{|\mathcal{GT}|}
		\end{eqnarray*}


	\item High \introduce{precision}:\\
		A good pattern should also be precise.
		For each individual ground truth pair $(s_i, t_i) \in \mathcal{GT}$ we can define the precision as:
		\begin{eqnarray*}
			\text{precision}_{gp}((s_i, t_i)) & = & \frac{|\{t_i | t_i\in \text{prediction}_{gp}(s_i)\}|}{|\text{prediction}_{gp}(s_i)|} \\
			\
		\end{eqnarray*}
		The target $t_i$ should be in the returned result list and if possible nothing else.
		In other words, we are not searching for patterns that return thousands of potentially wrong target for a given source.
		Over all ground truth pairs, we can define the average precision for $gp$ via the inverse of the average result lengths:
		\begin{eqnarray*}
			\text{avg result length}_{gp} & = & \mathop{\mathrm{avg}}_{(s_i, t_i)} \; |\text{prediction}_{gp}(s_i)| \\
			\text{precision}_{gp} & = & (\text{avg result length}_{gp})^{-1}
		\end{eqnarray*}

	\item High \introduce{gain}:\\
		A pattern discovered in run $r$ is better if it covers those ground truth pairs $gtp \in \mathcal{GT}$ that aren't covered with high precisions in previous runs ($gp' \in run_q)$ already:
		\begin{eqnarray*}
			\text{gain}_{run_r, gp} = \sum_{gtp} \max\{0, \text{precision}_{gp}(gtp) - \max_{\forall q < r: gp' \in run_q} \text{precision}_{gp'}(gtp))\}
		\end{eqnarray*}
		Similarly, the potentially remaining gain can be computed as:
		$$\text{remains}_{run_r} = \sum_{gtp \in \mathcal{GT}}(1 - \max_{\forall q < r: gp' \in run_q} \text{precision}_{gp'}(gtp))$$

	\item No \introduce{over-fitting}:\\
		While precision is to be maximised, a good pattern should not \introduce{over-fit} to a single source or target from the training input.

	\item Short \introduce{pattern length} and low \introduce{variable count}:\\
		If all other considerations are similar, then a shorter pattern or one with less variables is preferable.

		Note, that this is a low priority dimension.
		A good pattern is not restricted to a shortest path between \source and \target. 
		Good patterns can be longer and can have edges off the connecting path (e.g., see $gp_2$ in Section~\ref{sec:intro}).

	\item Low execution \introduce{time} \& \introduce{timeout}:\\
		Last but not least, to have any practical relevance, good patterns should be executable in a short \introduce{time}.
		Especially during the training phase, in which many queries are performed that take too long, we need to make sure to early terminate such queries on both, the graph pattern learner and the endpoint (cf. Section~\ref{sec:real_world_considerations}).
		In case the query was aborted due to a \introduce{timeout} and only a partial result obtained, it should not be trusted.

	\end{itemize}

	Based on these considerations, we define the \introduce{fitness} of an individual graph pattern as a tuple of real numbers with the following optimization directions.
	When comparing the fitness of two patterns, the fitness tuples for now are compared lexicographically.

	\begin{enumerate}
		\small
		\item \textbf{Remains} (max):
			Remaining precision sum $\text{remains}_{run_r}$ in the current run $r$ (see Section~\ref{sec:coverage}).
			Patterns found in earlier runs are considered better.
		\item \textbf{Score} (max):
			A derived attribute combining gain with a configurable multiplicative punishment for over-fitting patterns.
		\item \textbf{Gain} (max):
			The summed gained precision over the remains of the current run $r$: $\text{gain}_{run_r, gp}$.
			In case of timeouts or incomplete patterns the gain is set to 0.
		\item \textbf{$F_1$-measure} (max):
			$F_1$-measure for precision and recall of this pattern.
		\item \textbf{Average Result Lengths} (min):
			$\text{avg result length}_{gp}$.
		\item \textbf{Recall (Ground Truth Matches)} (max):
			$\text{gt\_matches}_{gp}$.
		\item \textbf{Pattern Length} (min):
			The number of triples this pattern contains.
		\item \textbf{Pattern Variables} (min):
			The number of variables this pattern contains.
		\item \textbf{Timeout} (min):
			Punishment term for timeouts (0.5 for a soft and 1.0 for a hard timeout) (see Section~\ref{sec:real_world_considerations} and gain).
		\item \textbf{Query Time} (min):
			The evaluation time in seconds.
			This is particularly relevant since it hints at the real complexity of the pattern.
			I.e., a pattern may objectively have a small number of triples and variables, but its evaluation could involve a large portion of the dataset.
	\end{enumerate}

\subsection{Initial Population}\label{sec:init_population}
	In order to start any evolutionary algorithm an initial population needs to be generated.
	The main objective of the first population is to form a starting point from which the whole search space is reachable via mutations and mating over the generations.
	While the initial population is not meant to immediately solve the whole problem, a poorly chosen initial population results in a lot of wasted computation time.

	The starting point of our algorithm are single triple SPARQL BGP queries, consisting only of variables with at least a \source and \target variable, e.g.:
	\pattern{?source ?p1 ?v1.}
	While having a small chance of survival (direct evaluation would typically yield bad fitness), such patterns can re-combine (see mating in Section~\ref{sec:mating}) with other patterns to form good and complete patterns in later generations.

	For prediction capabilities, we are searching graph patterns which connect \source and \target, our algorithm mostly fills the initial population with path patterns of varying lengths $l$ between \source and \target.
	Initially such a path pattern purely consists of variables and is directed from source to target:
	\pattern{\source $?p_1$ $?n_1$. $\ldots$ $?n_i$ $?p_{i + 1}$ $?n_{i + 1}$ . $\ldots$ $?n_{l-1}$ $?p_l$ \target.}
	For example a pattern of desired length of $l=3$ looks like this:
	\pattern{?source ?p1 ?n1. ?n1 ?p2 ?n2. ?n2 ?p3 ?target.}
	As longer patterns are less desirable, they are generated with a lower probability. 
	Furthermore, we randomly flip each edge of the generated patterns, in order to explore edges in any direction.


	In order to reduce the high complexity and noise introduced by patterns only consisting of variables, we built in a high chance to immediately subject them to the fix variable mutation (see Section~\ref{sec:mutation}).

\subsection{Mating}\label{sec:mating}
	In each generation there is a configurable chance for two patterns to mate in order to exchange information.
	In our algorithm this is implemented in a way that mating always creates two children, having the benefit of keeping the amount of individuals the same.
	Each child has a dominant and a recessive parent.
	The child will contain all triples that occur in both parents.
	Additionally, there is a high chance to select each of the remaining triples from the dominant parent and a low chance to select each of the remaining triples from the recessive parent.
	By this the children have the same expected length as their parents.

	Furthermore, as variables from the recessive parent could accidentally match variables already being in the child, and this can be beneficial or not, we add a 50~\% chance to rename such variables before adding the triples.

\subsection{Mutation}\label{sec:mutation}
	Besides mating, which exchanges information between two individuals, information can also be gained by mutation.
	Each individual in a population has a configurable chance to mutate by the following (non exclusive) mutation strategies.
	Currently, all but one of the mutation operations can be performed on the pattern itself (local) without issuing any SPARQL queries.
	The mutation operations also have different effects on the pattern itself (grow, shrink) and on its result size (harden, loosen).

	\begin{itemize}
		\small
		\item \textbf{introduce var} select a component (node or edge) and convert it into a variable (loosen) (local)
		\item \textbf{split var} select a variable and randomly split it into 2 vars (grow, loosen) (local)
		\item \textbf{merge var} select 2 variables and merge them (shrink, harden) (local)
		\item \textbf{del triple} delete a triple statement (shrink, loosen) (local)
		\item \textbf{expand node} select a node, and add a triple from its expansion (grow, harden) (local for now)
		\item \textbf{add edge} select 2 nodes, add an edge in between if available (grow, harden) (local for now)
		\item \textbf{increase dist} increase distance between source and target by moving one a hop away (grow) (local)
		\item \textbf{simplify pattern} simplify the pattern, deleting unnecessary triples (shrink) (local) (cf.\ Section~\ref{sec:pattern_simplification})
		\item \textbf{fix var} select a variable and instantiate it with an IRI, BNode or Literal that can take its place (harden) (SPARQL) (see below)
	\end{itemize}

	In a single generation sequential mutation (by different strategies in the order as above) is possible.

	We can generally say that introducing a variable loosens a pattern and fixing a variable hardens it.
	Patterns which are too loose will generate a lot of candidates and take a long time to evaluate.
	Patterns which are too hard will generate too few solutions, if any at all.
	Very big patterns, even though very specific can also exceed reasonable query and evaluation times.

	\subsubsection{Fix Var Mutation}\label{sec:mutate_fix_var}
	Unlike the other mutations, the fix var mutation is the only one which makes use of the underlying dataset via the SPARQL endpoint $G$, in order to instantiate variables with an IRI, BNode or Literal.
	As it is one of the most important mutations and also because performing SPARQL queries is expensive, it can immediately return several mutated children.

	For a given pattern $gp$ we randomly select one of its variables \sparql{?v} (excluding \source and \target).
	Additionally, we sample up to a defined number of source-target-pairs from the ground truth which are not well covered yet (high potential gain).
	For each of these sampled pairs $(s_s, t_s)$ we issue a SPARQL Select query of the form:
	\pattern{
	SELECT distinct ?v \{
		VALUES (?source ?target) \{ ($s_s$, $t_s$) \}
		$gp$
	\}
	}
	We collect the possible instantiations for \sparql{?v}, count them over all queries and randomly select (with probabilities according to their frequencies) up to a defined number of them.
	Each of the selected instantiations forms a separate child by replacing \sparql{?v} in the current pattern.

\subsection{Selection and Keeping the Population Healthy}\label{sec:population_control}
	After each generation the next generation is formed by the surviving (fittest) individuals from $n$ tournaments of $k$ randomly sampled individuals from the previous generation.


	We also employ two techniques, to counter population degeneration in local maxima and make our algorithm robust (even against non-optimal parameters):
	\begin{itemize}
		\small
		\item In each generation we re-introduce a small number of newly generated initial population patterns (see Section~\ref{sec:init_population}).
		\item Each generation updates a hall of fame, which will preserve the best patterns ever encountered over the generations.
		In each generation a small number of the best of these all-time best patterns is re-introduced.
	\end{itemize}

\subsection{Real World Considerations}\label{sec:real_world_considerations}
	In the following, we will briefly discuss practical problems that we encountered and necessary optimizations we used to overcome them.
	We implemented our graph pattern learner with the help of the DEAP (Distributed Evolutionary Algorithms in Python) framework~\cite{Fortin2012DEAP}.




	\subsubsection{Batch Queries}
	The single most important optimization of our algorithm lies in the reduction of the amount of issued queries by using batch queries.
	This mostly applies to the queries for fitness evaluation (Section~\ref{sec:fitness}).
	It is a lot more efficient to run several sub-queries in one big query and to only transport the ground truth pairs to the endpoint once (via \sparql{VALUES}), than to ask for each result separately.



	\subsubsection{Timeouts \& Limits}
	Another mandatory optimization involves the use of timeouts and limits for all queries, even if they usually only return very few results in a short time.
	We found that a few run-away queries can quickly lead to congestion of the whole endpoint and block much simpler queries.

	Timeouts are also especially useful as a reliable proxy to exclude too complicated graph patterns.
	Even seemingly simple patterns can take a very long time to evaluate based on the underlying dataset and its distribution.

	\subsubsection{Fit To Live Filter}
	Apart from timeouts we use a filter which checks if mutants and children are actually desirable (e.g., length and variable count in boundaries, pattern is complete and connected), meaning fit to live, before evaluating them.
	If not, the respective parent takes their place in the new population, allowing for a much larger part of the population to be viable.


	\subsubsection{Parallelization, Caching, Query Canonicalization and Noise}
	Two other crucial optimizations to reduce the overall run-time of the algorithm are parallelization and client side caching.
	Evolutionary algorithms are easy to parallelize via parallel evaluation of all individuals, but in our case the SPARQL endpoint quickly becomes the bottleneck.
	Ignoring the limits of the queried endpoint will resemble a denial of service attack.
	For most of our experiments we hence use an internal LOD cache with exclusive access for our learning algorithm.
	In case the algorithm is run against public endpoints we suggest to only use a single thread in order not to disturb their service (fair use).

	Client side caching further helps to reduce the time spent on evaluating graph patterns, by only evaluating them once, should the same pattern be generated by different sequences of mutation and mating operations.
	To identify equivalent patterns despite different syntactic surface forms, we had to solve SPARQL BGP canonicalization (finding a canonical graph labelling).
	We were able to reduce the problem to RDF graph canonicalization and achieve good practical run-times with RGDA1~\cite{McCusker2015RGDA1}.


	In the context of caching, one other important finding is that many SPARQL endpoints (especially the widely used OpenLink Virtuoso) often return incomplete and thereby non-deter\-mi\-nis\-tic results by default.
	Unlike many other search algorithms, an evolutionary algorithm has the benefit that it can cope well with such non-determinism.
	Hence, when caching is used, it is helpful to reduce, but not completely remove redundant queries.

	\subsubsection{Pattern Simplification}\label{sec:pattern_simplification}
	Last but not least, as our algorithm can create patterns that are unnecessarily complex, it is useful to simplify them.
	We developed a pattern simplification algorithm, which given a complicated graph pattern $gp_c$ finds a minimal equivalent pattern $gp_s$ with the same result set wrt. the \source and \target variables.
	The simplification algorithm removes unnecessary edges, such as redundant parallel variable edges, edges between and behind fixed nodes and unrestricting leaf branches.


\section{Visualisation}\label{sec:visualisation}
	After presenting the main components of our evolutionary algorithm in the previous section, we will now briefly present an interactive visualisation\footnote{Also available at \url{https://w3id.org/associations}.}.
	As the learning of our evolutionary algorithm can produce many graph patterns, the visualisation allows to quickly get an overview of the resulting patterns in different stages of the algorithm.

	\begin{figure}[bt]
		\centering
		\includegraphics[width=.49\textwidth]{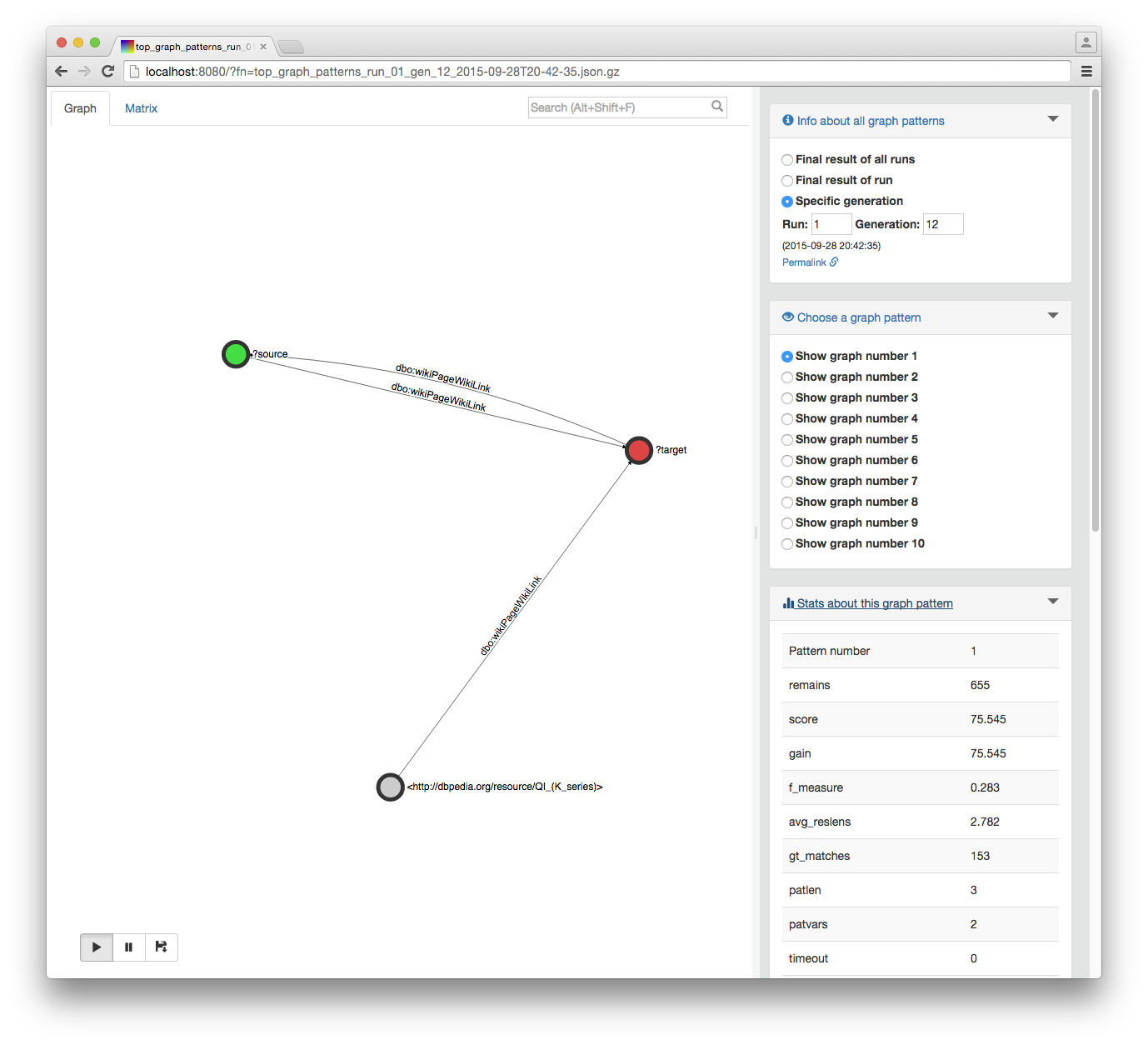}
		\includegraphics[width=.49\textwidth]{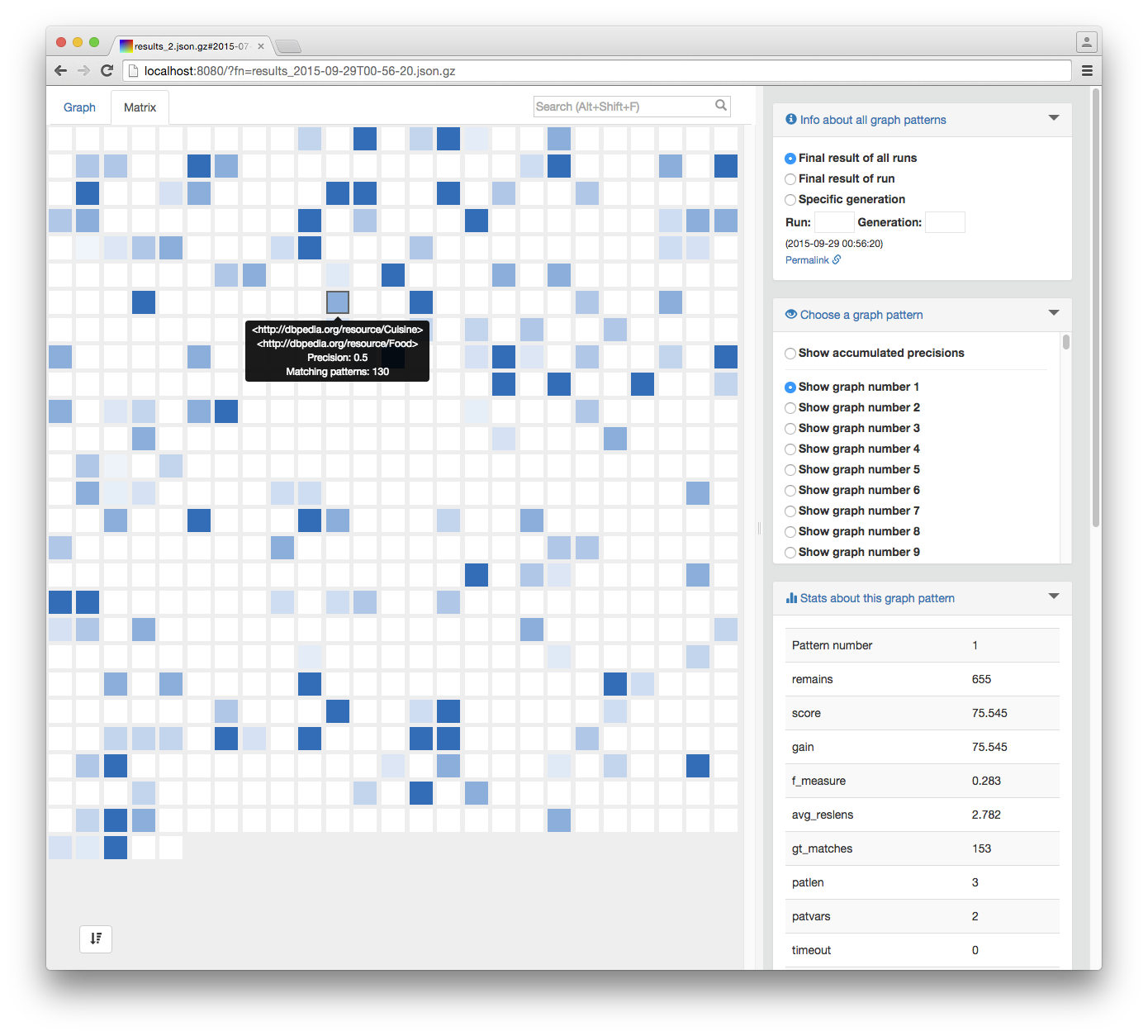}
		\caption{Visualisation of graph pattern 1 from run 1, generation 12 (left) and
			the precision vector over all training ground truth pairs of graph pattern 1 (right).
			Each block represents a $(?source, ?target)$ pair from the ground truth training set.
			The darker its colour the higher the precision for the ground truth pair.
		}
		\label{fig:visualisation_gp_results}
		\label{fig:visualisation_coverage}
	\end{figure}

	Figure~\ref{fig:visualisation_gp_results}~(left) shows a screen shot of the visualisation of a single learned graph pattern.
	In the sidebar the user can select between individual generations, the results of a whole run or the overall results (default) to inspect the outcomes at various stages of the algorithm.
	Afterwards, the individual result graph patterns can be selected.
	Below these selection options the user can inspect statistics about the selected graph pattern including its fitness, a list of matching training ground truth pairs and the corresponding SPARQL SELECT query for the pattern.
	Links are provided to perform live queries on the SPARQL endpoint.

	At each of the stages, the user can also get an overview of the precision coverage of a single pattern (as can be seen in Figure~\ref{fig:visualisation_coverage}~(right)) or the accumulated coverage over all patterns.

\section{Prediction}\label{sec:prediction}
	As already mentioned in the introduction, the learned patterns can be used to predict targets for a given source.
	The basic idea is to insert a given source $s_i$ in place of the \source variable in each of the learned patterns $gp \in gpl(\mathcal{GT}, G)$ and execute a SPARQL Select query over the \target variable (c.f., $\text{prediction}_{gp}(s_i)$ in Section~\ref{sec:intro}).

	\subsection{Query Reduction Technique}\label{sec:query_reduction}

		While interesting for manual exploration, for practical prediction purposes the amount of learned graph patterns can easily become too large by discovering many very similar patterns that are only differing in minor aspects.

		One realisation from visualising the resulting patterns $gp$, is that we can use their precision vectors wrt. the ground truth pairs to cluster graph patterns.
		The $i$-th component of $\vec{pv}_{gp}$ is defined by the precision value corresponding to the $i$-th ground truth source-target-pair $stp_i \in \mathcal{GT}$:
		$$pv_{gp,i} = \text{precision}_{gp}(stp_i)$$

		We employ several standard clustering algorithms on $\vec{pv}_{gp}$ and select the best patterns $\text{cluster}(gpl(\mathcal{GT}, G))$ in each cluster as representatives to reduce the amount of queries. 
		By default our algorithm applies all of these clustering techniques and then selects the one which minimises the precision loss at the desired number of queries to be performed during prediction.

		In our tests we could observe, that clustering (e.g., with hierarchical scaled euclidean ward clustering) allows us to reduce the number of performed SPARQL queries to 100 for all practical purposes with a precision loss of less than $1\%$.

	\subsection{Fusion Variants}\label{sec:fusion}
		When used for prediction, each graph pattern $gp$ 
		creates an unordered list of possible target nodes $t_j \in \text{prediction}_{gp}(s_i)$ for an inserted source node $s_i$.
		We evaluated the following fusion strategies to combine and rank the returned target candidates $t_j$ (higher fusion value means lower rank):
		\vspace{-1.5ex}
		\begin{itemize}
		\small
		\itemsep-.5ex
			\item \textbf{target occurrences}: a simple occurrence count of each of the targets over all graph patterns.
			\item \textbf{scores}: sum of all graph pattern scores (from the graph pattern's fitness) for each returned target.
			\item \textbf{f-measures}: sum of all graph pattern $F_1$-measures (from the graph pattern's fitness) for each returned target.
			\item \textbf{gp precisions}: sum of all graph pattern precisions (from the graph pattern's fitness) for each returned target.
			\item \textbf{precisions}: sum of the actual precisions per graph pattern in this prediction.
		\end{itemize}
		By default our algorithm will calculate them all, allowing the user to pick the best performing fusion strategy for their use-case.

\section{Evaluation}\label{sec:eval}
	In order to evaluate our graph pattern learner, we performed several experiments which we will describe in the following.

	We ran our experiments against a local Virtuoso 7.2 SPARQL endpoint containing over $7.9$ G triples, from many central datasets\footnote{Most notably: DBpedia 2015-04 (en, de), Freebase, Yago, Wikidata, GeoNames, DBLP, Wordnet and BabelNet.} of the LOD cloud, denoted as $G$ in the following.

	\subsection{Single Pattern Re-Identification}
	One of our claims is that our algorithm can learn good SPARQL queries for a relation $\mathcal{R}$ represented by a set of ground truth source-target-pairs $\mathcal{GT}$.
	In order to evaluate this, we started with simple relations such as ``given a capital $s$ return its country $t$'' (see $\mathcal{R}_{cc}$ in Section~\ref{sec:intro}).
	For each $\mathcal{R}$, we used a generating SPARQL query $gp_g$ (such as $gp_2$ from Section~\ref{sec:intro}) to generate $\mathcal{GT} \subset \text{SELECT}(gp_g)$, then executed our graph pattern learner $gpl(\mathcal{GT}, G)$ and checked if $gp_g$ was in the resulting patterns:
	$$gp_g \mathop{\in}^{?} gpl(\mathcal{GT}, G)$$

	The result of these experiments is that our algorithm is able to re-identify such simple, readily modelled relations $\mathcal{R}$ in $100\%$ of our test cases (typically within the first run, so the first 3 minutes).
	While this might sound astonishing, it is merely a test that our algorithm can perform the simplest of its tasks:
	If there is a single SPARQL BGP pattern $gp$ that models the whole training list $\mathcal{GT}$ in $G$, then our algorithm is quickly able to find it via the fix var mutation in Section~\ref{sec:mutate_fix_var}.
	Due to the page limit, we omit further details and instead turn to a more complex relation in the next section.

	\subsection{Learning Patterns for Human Associations from DBpedia}
	Two additional claims are that our algorithm can learn a set of patterns, which cover a complex relation $\mathcal{R}$ that is not readily modelled in $G$, and that we can use the resulting patterns for prediction.
	Hence, in the following we focus on one such complex relation $\mathcal{R}_{ha}$: human associations.
	We will present some of the identified patterns and then evaluate the prediction quality.

	\subsubsection{Dataset}
		Human associations are an important part of our thinking process.
		An \introduce{association} is the mental connection between two ideas: a \introduce{stimulus} (e.g., \stimulus{pupil}) and a \introduce{response} (e.g., \response{eye}).
		We call such associations \introduce{strong associations} if more than 20~\% of people agree on the response.



	In the following, we focus on a dataset of 727 strong human associations (corresponding to $\sim 25.5$ K raw associations) from the Edinburgh Associative Thesaurus~\cite{Kiss1973EAT} that we previously already mapped to DBpedia Entities~\cite{Hees2016EAT}.
	The dataset contains stimulus-response-pairs such as \semassociation{\dbr{Pupil}}{\dbr{Eye}}, \semassociation{\dbr{Stanza}}{\dbr{Poetry}} and \semassociation{\dbr{Paris}}{\dbr{France}}.\footnote{The full dataset is available at \url{https://w3id.org/associations}.}




	We randomly split our 727 ground truth pairs into a training set $\mathcal{GT}_\text{train}$ of 655 and a test set $\mathcal{GT}_\text{test}$ of 72 pairs (10~\% random split).
	All training, visualising and development has been performed on the training set in order to reduce the chance of over-fitting our algorithm to our ground truth.

	\subsubsection{Basic Statistics}
		%
		We ran the algorithm ($gpl(\mathcal{GT}_\text{train}, G)$) on $G$ with a population size of 200, a maximum of 20 generations each in a maximum of 64 runs.
		The first 5 runs of our algorithm are typically completed within 3, 6, 9, 13 and 15 minutes.
		In the first couple of minutes all of the very simple patterns that model a considerable fraction of the training set's pairs are found.
		With the mentioned settings the algorithm will terminate after around 3 hours.
		It finds roughly 530 graph patterns with a score > 2 (cf.\ Section~\ref{sec:fitness}).

	\subsubsection{Notable Learned Graph Patterns}
		Due to the page limit, we will briefly mention only 3 notable patterns from the resulting learned patterns in this paper.
		We invite the reader to explore the full results online\footnote{\url{https://w3id.org/associations}} with the interactive visualisation presented in Section~\ref{sec:visualisation}.
		The three notable patterns we want to present here are:
		\pattern{?source \gold{hypernym} ?target}
		\pattern{?source \dbo{wikiPageWikiLink} ?target. ?target \dbo{wikiPageWikiLink} ?source}
		\pattern{?source \dbo{wikiPageWikiLink} ?target. ?v0 \skos{exactMatch} ?v1. ?v1 \dbprop{industry} ?target}

		The first two are intuitively understandable patterns which typically are amongst the top patterns.
		The first one shows that human associations often seem to be represented via \gold{hypernym} in DBpedia (the response is often a hypernym (broader term) for the stimulus).
		The second one shows that associations often correspond to bidirectionally linked Wikipedia articles.
		The third pattern represents a whole class of intra-dataset learning by making use of a connection of the \target to BabelNet's \skos{exactMatch}.

	\subsubsection{Prediction \& Fusion Strategies Evaluation}\label{sec:eval:fusion}
		As human associations are not readily modelled in DBpedia, it is difficult to assess the quality of the learned patterns $gp$ directly.
		Hence, we evaluate the quality indirectly via their prediction quality on the test-set $\mathcal{GT}_\text{test}$.

		For each of the $(s_t, t_t) \in \mathcal{GT}_\text{test}$ we generate a ranked target list $rtpl_{s_t} = [tp_1, \ldots, tp_n]$ of target predictions $tp_i$.
		The list is the result of one of the fusion variants (cf.\ Section~\ref{sec:fusion}) after clustering (cf.\ Section~\ref{sec:query_reduction}).
		For evaluation, we can then check the rank $r_t$ of $t_t$ in $rtpl_{s_t}$ (lower ranks are better).
		If $t_t \notin rtpl_{s_t}$, we set $r_t = \infty$.

		An example of a ranked target prediction list (for the fusion method \emph{precisions}) for source $s_t=$\dbr{Sled} is the ranked list: $rtpl_\dbr{Sled} = $ [\dbr{Snow}, \dbr{Christmas}, \dbr{Deer}, \dbr{Kite}, \dbr{Transport}, \dbr{Donkey}, \dbr{Ice}, \dbr{Ox}, \dbr{Obelisk}, \dbr{Santa_Claus}].
		In this case the ground truth target $t_t=$\dbr{Snow} is at rank $r_t = 1$.
		As we can see most of the results are relevant as associations to humans.
		Nevertheless, for the purpose of our evaluation, we will only consider the single $t_t$ corresponding to a $s_t$ as relevant and all other $tp_i$ as irrelevant.

\begin{figure}[tb]
			\centering
			\includegraphics[width=.8\textwidth]{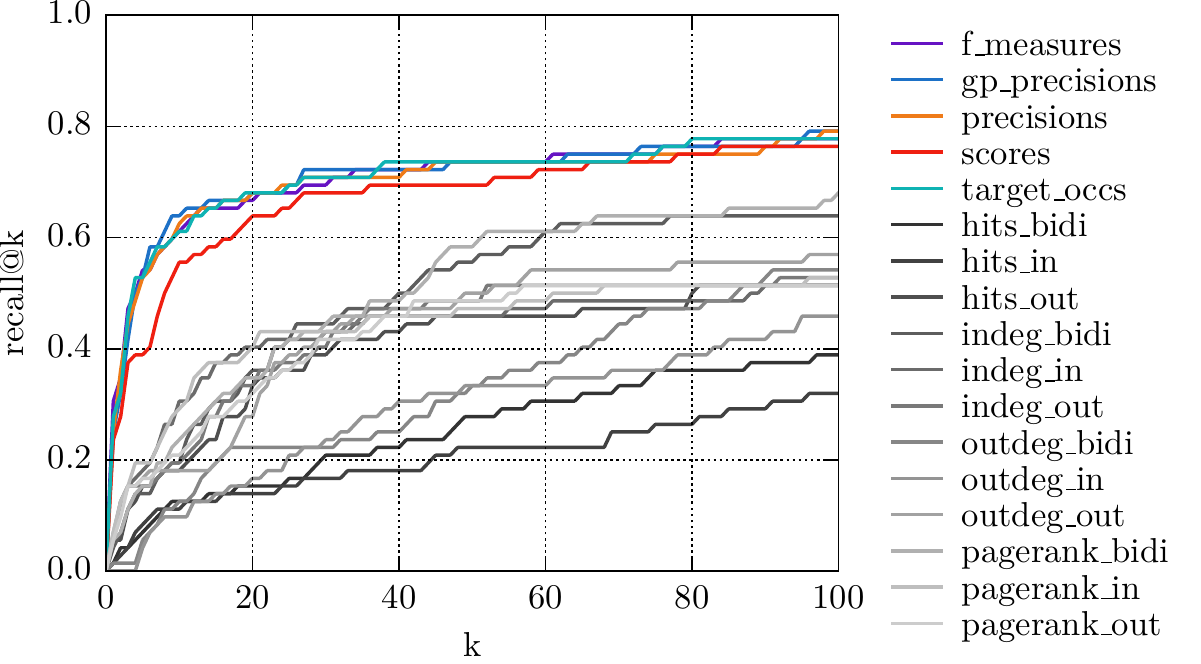}
			\caption{Recall$@$k over the different fusion variants and against baselines.}
			\label{fig:recall}
		\end{figure}

		Based on the ranked result lists, we can calculate the Recall@k\footnote{We don't provide Precision@k, as it degenerates to $\text{Recall@k}/k$ due to the fact that we only have 1 relevant target per result of any $(s_t, t_t)$.}, Mean Average Precision (MAP) and Normalised Discounted Cumulative Gain of the various fusion variants over the whole test set $\mathcal{GT}_\text{test}$, as can be seen in Table~\ref{tbl:eval} and Figure~\ref{fig:recall}.

		We also calculate these metrics for several baselines, which try to predict the target nodes from the 1-neighbourhood (bidirectionally, incoming or outgoing) by selecting the neighbour with the highest PageRank, HITS score, in- and out-degree~\cite{ReddyKnuthSack2014DBpediaGraphMeasures,Thalhammer2016DBpediaPagerank}.
		As can be seen, all our fusion strategies significantly outperform the baselines.

		\begin{table}[bth]
		\newcommand{\bb}[1]{\textbf{#1}}
		\newcommand{\rc}[1]{\tiny Recall@#1}
		\newcommand{\mc}[3]{\multicolumn{#1}{#2}{#3}}
		\centering \scriptsize
		\caption{Recall@k, MAP and NDCG for our fusion variants and against baselines.}
		\label{tbl:eval}
		\begin{tabular}{l|cccccccc}
			                 &    \rc{1}  &    \rc{2}  &    \rc{3}  &    \rc{4}  &    \rc{5}  &   \rc{10}  &     MAP    &      NDCG  \\
			\hline
			       outdeg in &     0.000  &     0.000  &     0.000  &     0.000  &     0.042  &     0.097  &     0.029  &     0.105  \\
			      outdeg out & \bb{0.069} & \bb{0.125} & \bb{0.153} &     0.153  &     0.167  &     0.181  &     0.126  &     0.209  \\
			     outdeg bidi &     0.014  &     0.014  &     0.014  &     0.014  &     0.056  &     0.125  &     0.045  &     0.131  \\
			[1ex]
			        indeg in &     0.056  &     0.111  & \bb{0.153} &     0.167  &     0.181  & \bb{0.306} &     0.129  &     0.207  \\
			       indeg out &     0.056  & \bb{0.125} & \bb{0.153} &     0.153  &     0.153  &     0.194  &     0.121  &     0.200  \\
			      indeg bidi &     0.042  &     0.069  &     0.111  &     0.139  &     0.139  &     0.194  &     0.104  &     0.205  \\
			[1ex]
			     pagerank in & \bb{0.069} & \bb{0.125} & \bb{0.153} & \bb{0.194} & \bb{0.194} &     0.292  & \bb{0.140} & \bb{0.219} \\
			    pagerank out &     0.056  &     0.097  & \bb{0.153} &     0.153  &     0.167  &     0.208  &     0.117  &     0.195  \\
			   pagerank bidi &     0.056  &     0.069  &     0.111  &     0.139  &     0.153  &     0.236  &     0.113  & \bb{0.219} \\
			[1ex]
			         hits in &     0.014  &     0.028  &     0.042  &     0.069  &     0.083  &     0.111  &     0.046  &     0.095  \\
			        hits out &     0.056  &     0.056  &     0.111  &     0.125  &     0.153  &     0.181  &     0.102  &     0.181  \\
			       hits bidi &     0.014  &     0.042  &     0.042  &     0.056  &     0.069  &     0.125  &     0.050  &     0.110  \\
			[.5ex]\hline
			          scores &     0.236  &     0.278  &     0.375  &     0.389  &     0.389  &     0.556  &     0.323  &     0.413  \\
			   gp precisions &     0.250  &     0.319  &     0.417  &     0.500  &     0.528  & \bb{0.639} &     0.365  &     0.457  \\
			      precisions &     0.250  & \bb{0.361} &     0.444  &     0.486  &     0.528  &     0.625  &     0.371  &     0.460  \\
			     target occs &     0.278  &     0.319  &     0.458  & \bb{0.528} &     0.528  &     0.611  &     0.381  &     0.466  \\
			      f measures & \bb{0.306} &     0.347  & \bb{0.472} &     0.500  & \bb{0.542} &     0.611  & \bb{0.399} & \bb{0.479} \\
		\end{tabular}
		\end{table}

\section{Conclusion \& Outlook}\label{sec:conclusion}\label{sec:future_work}
	In this paper we presented an evolutionary graph pattern learner.
	The algorithm can successfully learn a set of patterns for a given list of source-target-pairs from a SPARQL endpoint.
	The learned patterns can be used to predict targets for a given source.

	We use our algorithm to identify patterns in DBpedia for a dataset of human associations.
	The prediction quality of the learned patterns after fusion reaches a Recall$@$10 of 63.9~\% and MAP of 39.9~\%, and significantly outperforms PageRank, HITS and degree based baselines.

	The algorithm, the used datasets and the interactive visualisation of the results are available online\footnote{\url{https://w3id.org/associations}}.

	In the future, we plan to enhance our algorithm to support Literals in the input source-target-pairs, which will allow us to learn patterns directly from lists of textual inputs. 
	Further, we are investigating mutations, for example to introduce \sparql{FILTER} constraints. 
	We also plan to investigate the effects of including negative samples (currently we only use positive samples and treat everything else as negative).

	Additionally, we plan to employ more advanced late fusion techniques, in order to learn when to trust the prediction of which pattern.
	As this idea is conceptually close to interpreting the learned patterns as a feature vector (with understandable and executable patterns to generate target candidates), we plan to investigate combinations of our algorithm with approaches that learn vector space representations from knowledge graphs.

	This work was supported by the University of Kaiserslautern CS PhD scholarship program and the BMBF project MOM (Grant 01IW15002).


\end{document}